\documentclass{article}

\usepackage[preprint]{neurips_2023}

\usepackage[utf8]{inputenc} % allow utf-8 input
\usepackage[T1]{fontenc}    % use 8-bit T1 fonts
\usepackage{hyperref}       % hyperlinks
\usepackage{url}            % simple URL typesetting
\usepackage{booktabs}       % professional-quality tables
\usepackage{amsfonts}       % blackboard math symbols
\usepackage{nicefrac}       % compact symbols for 1/2, etc.
\usepackage{microtype}      % microtypography
\usepackage{xcolor}         % colors
\usepackage{amsmath}
\usepackage{multirow}
\usepackage{graphicx}
\usepackage{gensymb}
\usepackage{courier}
\usepackage{natbib}
\usepackage{textcomp}
\usepackage{makecell}
\usepackage{algorithm}%
\usepackage{algorithmicx}%
\usepackage{algpseudocode}
\usepackage{tablefootnote}
\setcitestyle{numbers,square}

\title{Learning the Plasticity: Plasticity-Driven Learning Framework in Spiking Neural Networks}

\author{
	{Guobin Shen$^{1, 3}$\footnotemark[1], \ Dongcheng Zhao$^{1}$\footnotemark[1], \ Yiting Dong$^{1, 3}$,} \\ 
    \textbf{\ Yang Li$^{1, 4}$, Feifei Zhao $^{1}$ \ Yi Zeng$^{1, 2, 3, 4}$\footnotemark[2]} \\
	  $^1$ Brain-inspired Cognitive Intelligence Lab, Institute of Automation, Chinese Academy of Sciences\\ 
      $^2$ Center for Excellence in Brain Science and Intelligence Technology, CAS\\
      $^3$ School of Future Technology, University of Chinese Academy of Sciences \\
      $^4$ School of Artificial Intelligence, University of Chinese Academy of Sciences \\
	\texttt{\{shenguobin2021, zhaodongcheng2016, dongyiting2020, } \\ 
    \texttt{liyang2019, zhaofeifei2014, yi.zeng\}@ia.ac.cn}
}

\begin{document}

\maketitle

\renewcommand{\thefootnote}{\fnsymbol{footnote}}
\footnotetext[1]{Equal contribution.}
\footnotetext[2]{Corresponding Author.}
\renewcommand{\thefootnote}{\arabic{footnote}}

% TL;DR 在仅优化随机权重脉冲神经元的异质性参数时, 我们发现这样的异质性SNNs能够在强化学习, 工作记忆等任务上实现比同质性SNNs更好的性能, 并发现膜时间常数对于异质性SNNs至关重要, 并具有和生物神经元相似的分布. 

% Optimizing only heterogeneity parameters of random weighted SNN can improve performance in reinforcement learning and working memory. Membrane time constants are critical for heterogenous SNNs and exhibit a distribution similar to biological neurons.

% 人类大脑的自然进化产生了多种突触可塑性, 它们可以动态地变化以适应不断变化的世界。突触可塑性的演化激发了我们去探索生物学上合理的尖峰神经网络（SNNs）优化和学习算法。当前的神经网络依赖于突触权重的直接训练, 这一过程最终导致了固定的连接, 从而阻碍了它们适应动态的现实世界环境的能力。为了改善这种情况, 我们引入了元可塑性的应用--一种涉及学习可塑性规则的复杂机制, 而不是参与突触权重的直接修改。元可塑性动态能够动态地结合不同的可塑性规则, 有效地增强了工作记忆、多任务泛化和适应, 并揭示了各种可塑性与认知功能之间的潜在关联。通过将元可塑性整合到脉冲神经网络中, 我们展示了人工智能系统中增强的适应性和认知功能, 并从计算角度揭示了脑的学习机制, 是神经科学与人工智能深度交叉的一步。

\begin{abstract}
    % 人类大脑的进化导致了复杂的突触可塑性的发展，从而能够动态适应不断变化的世界。 这一进展激发了我们对尖峰神经网络（SNN）新范式的探索：以可塑性为中心的学习框架。 这种范式与传统的神经网络模型不同，传统的神经网络模型主要关注突触权重的直接训练，导致静态连接限制了动态环境中的适应性。 相反，我们的方法深入研究突触行为的核心，优先考虑可塑性规则本身的学习。 这种重点从权重调整到掌握突触变化的复杂性的转变为神经网络的进化和适应提供了更灵活和动态的途径。 我们以可塑性为中心的学习框架不仅适应了功能和稳态可塑性的现有概念，而且重新定义了它们，与生物学习的动态和适应性本质紧密结合。 这种重新定位增强了人工智能系统的关键认知能力，例如工作记忆和多任务处理能力，并在复杂的现实场景中表现出卓越的适应性。 此外，我们的框架揭示了各种形式的可塑性和认知功能之间的复杂关系，从而有助于更深入地理解大脑的学习机制。将这种突破性的以可塑性为中心的方法集成到 SNN 中标志着神经科学和人工智能融合的重大进步。 它为开发人工智能系统铺平了道路，这些系统不仅可以学习，而且可以适应不断变化的世界，就像人脑一样。
    The evolution of the human brain has led to the development of complex synaptic plasticity, enabling dynamic adaptation to a constantly evolving world. This progress inspires our exploration into a new paradigm for Spiking Neural Networks (SNNs): a Plasticity-Driven Learning Framework (PDLF). This paradigm diverges from traditional neural network models that primarily focus on direct training of synaptic weights, leading to static connections that limit adaptability in dynamic environments. Instead, our approach delves into the heart of synaptic behavior, prioritizing the learning of plasticity rules themselves. This shift in focus from weight adjustment to mastering the intricacies of synaptic change offers a more flexible and dynamic pathway for neural networks to evolve and adapt. Our PDLF does not merely adapt existing concepts of functional and Presynaptic-Dependent Plasticity but redefines them, aligning closely with the dynamic and adaptive nature of biological learning. This reorientation enhances key cognitive abilities in artificial intelligence systems, such as working memory and multitasking capabilities, and demonstrates superior adaptability in complex, real-world scenarios. Moreover, our framework sheds light on the intricate relationships between various forms of plasticity and cognitive functions, thereby contributing to a deeper understanding of the brain's learning mechanisms. Integrating this groundbreaking plasticity-centric approach in SNNs marks a significant advancement in the fusion of neuroscience and artificial intelligence. It paves the way for developing AI systems that not only learn but also adapt in an ever-changing world, much like the human brain.
\end{abstract}

\section{Introduction}\label{sec1}

% 突触可塑性的特征是突触随着时间的推移增强或减弱的能力，它支撑着人脑的学习和记忆。 这种自适应机制允许对不断变化的环境做出动态响应，体现在从分子到神经网络的各个层面~\cite{hebb2005organization,bienenstock1982theory, bi1998synaptic, turrigiano1999homeostatic}。 其重要性已得到广泛认可； 然而，突触可塑性作为人工神经网络中的优化算法的直接应用提出了显着的挑战。
Synaptic plasticity, characterized by the ability of synapses to strengthen or weaken over time, underpins learning and memory in the human brain. This adaptive mechanism allows for dynamic responses to an ever-evolving environment, manifesting across various levels from molecular to neural networks~\cite{hebb2005organization,bienenstock1982theory, bi1998synaptic, turrigiano1999homeostatic}. Its significance is widely recognized; however, the direct application of synaptic plasticity as an optimization algorithm in artificial neural networks presents notable challenges.

% 关键挑战之一来自生物学中观察到的学习规则的多样性，例如长期抑制 (LTD)、长期增强 (LTP) 和尖峰时序依赖性可塑性 (STDP)。 这些机制虽然有助于理解和实现学习和记忆等功能，但由于其复杂性和微妙性，直接应用于神经网络优化通常具有挑战性~\cite{diehlUnsupervisedLearningDigit2015, kheradpisheh2016bio, panda2016unsupervised, duan2023hebbian}。 受生物证据启发的传统建模方法难以捕捉神经系统的固有动态，并且需要精心的手工设计或与深度学习优化技术的集成。
One of the key challenges arises from the diversity of learning rules observed in biology, such as Long-Term Depression (LTD), Long-Term Potentiation (LTP), and Spike-Timing-Dependent Plasticity (STDP). These mechanisms, although instrumental in understanding and implementing functions like learning and memory, are often challenging to directly apply in neural network optimization due to their complexity and subtlety~\cite{diehlUnsupervisedLearningDigit2015, kheradpisheh2016bio, panda2016unsupervised, duan2023hebbian}. Traditional modeling methods, inspired by biological evidence, struggle to capture the inherent dynamism of neural systems and require meticulous hand-design or integration with deep learning optimization techniques.

% 尖峰神经网络 (SNN) 巧妙地模拟生物神经系统中的离散尖峰序列信息传输，对生物神经元的动力学进行复杂的建模。 SNN 中信息处理的内在事件驱动和实时性质赋予它们管理时间动态任务的增强能力，在这些方面优于传统的人工神经网络 (ANN)~\cite{maass1997networks}。 虽然反向传播已被确立为神经网络优化的基础技术~\cite{rumelhartLearningRepresentationsBackpropagating1986a, wuSpatioTemporalBackpropagationTraining2018}，但其在生物系统中的精确模拟仍然存在争议，引发了关于使用此类算法复制复杂生物任务的可行性的问题~\cite{lillicrap2020backpropagation} 。 优化 SNN 的另一种方法是利用生物学中的可塑性机制。 这种方法根据在生物系统中观察到的局部突触可塑性规则来复制复杂任务的挑战。 然而，通过各种学习规则来增强SNN的学习能力往往依赖于手动预设的协调，缺乏适应不同环境变化的灵活性~\cite{diehl2015unsupervised, dong2023unsupervised, zhao2020glsnn}，限制了该领域的发展。
Spiking Neural Networks (SNNs) adeptly emulate the discrete spike sequence information transmission found in biological nervous systems, intricately modeling the dynamics of biological neurons. The intrinsic event-driven and real-time nature of information processing in SNNs grants them an enhanced capability for managing tasks with temporal dynamics, outperforming traditional Artificial Neural Networks (ANNs) in these aspects~\cite{maass1997networks}. While backpropagation has been established as a foundational technique in neural network optimization~\cite{rumelhartLearningRepresentationsBackpropagating1986a, wuSpatioTemporalBackpropagationTraining2018}, its precise analog in biological systems remains controversial, raising questions about the feasibility of replicating complex biological tasks using such algorithms~\cite{lillicrap2020backpropagation}. Another way to optimize SNNs is to draw on plasticity mechanisms in biology. This approach replicates the challenges of complex tasks based on local synaptic plasticity rules observed in biological systems. However, enhancing the learning ability of SNN through various learning rules often relies on the coordination of manual presets and lacks the flexibility to adapt to changes in different environments~\cite{diehl2015unsupervised, dong2023unsupervised, zhao2020glsnn}, limiting the development of this field.

% 我们认为该领域的主要挑战源于神经网络设计中观察到的生物可塑性机制的严格应用。 对这些机制缺乏抽象和更深入的理解，导致模型无法充分利用突触变化的适应性和学习潜力。 为了克服这些限制，我们提出了一种创新方法，将重点从传统的突触权重调整转移到学习可塑性原理。 我们提倡对可塑性进行抽象和参数化建模，旨在学习和适应网络内的可塑性规则。 这种范式转变与生物学习的动态本质更加紧密地结合在一起，并为开发更强大和适应性更强的神经网络模型开辟了途径。
We argue that the main challenge in this field stems from the rigid application of observed biological plasticity mechanisms in neural network design. There is a lack of abstraction and deeper understanding of these mechanisms, leading to models that do not fully exploit the adaptability and learning potential of synaptic changes. To overcome these limitations, we propose an innovative approach that shifts the focus from traditional synaptic weight adjustments to learning the principles of plasticity. We advocate for an abstract and parametric modeling of plasticity, aiming to learn and adapt the rules of plasticity within the network. This paradigm shift aligns more closely with the dynamic nature of biological learning and opens avenues for developing more robust and adaptable neural network models.

% 我们的贡献可总结如下：
% \开始{枚举}
%      \item 我们提出生物可塑性的抽象和参数化建模，提供局部可塑性的更高水平的概括和总结，从而导致更灵活的形式和更好的泛化能力。
%      \item 我们引入了以可塑性为中心的学习框架，强调对可塑性规则的理解和应用，而不是传统的突触权重调整，使神经网络能够在动态环境中进化和适应。
%      这种方法有可能增强 SNN 在动态和现实场景中的泛化和多任务处理能力，提供持续学习和适应的平台，反映生物神经系统的非凡能力。
% \end{枚举}

Our contributions can be summarized as follows:
\begin{itemize}
    \item We propose abstract and parametric modeling of biological plasticity, providing a higher level of generalization and summary of local plasticity, leading to more flexible forms and better generalization capabilities.
    \item We introduce the Plasticity-Driven Learning Framework (PDLF), emphasizing the understanding and application of plasticity rules over traditional synaptic weight adjustments, allowing neural networks to evolve and adapt in dynamic environments.
    \item This approach potentially enhances the generalization and multitasking abilities of SNNs in dynamic and real-world scenarios, offering a platform for continuous learning and adaptation, mirroring the extraordinary capabilities of biological nervous systems.
\end{itemize}

\section{Results}\label{sec2}

\begin{figure}[t]
    \centering
    \includegraphics[width=\textwidth]{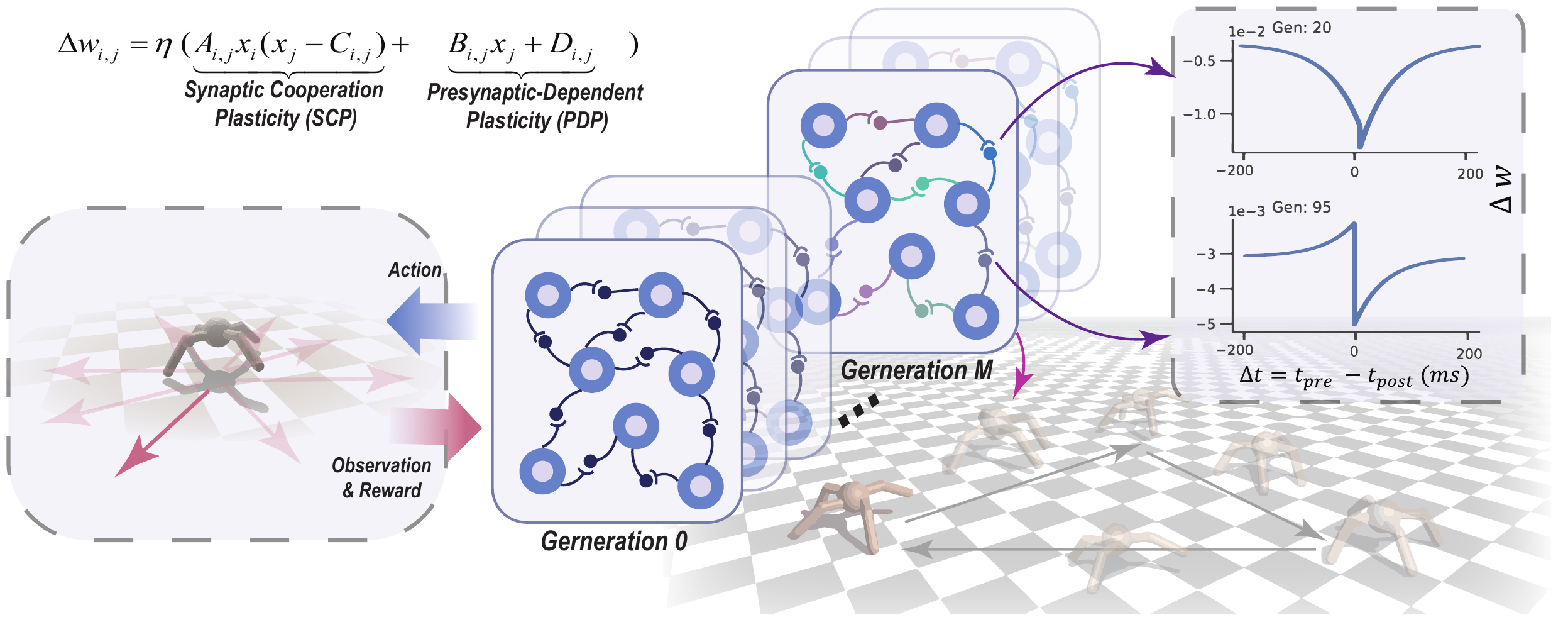}
    % 元可塑性的示意图. 顶部: 结合学习可塑性和稳态可塑性, 神经元能够实现多样化的, 异质的可塑性. 底部: 具有元可塑性的智能体通过学习可塑性而不是直接调整权重. 神经元之间能够形成不同的突触可塑性形式, 能够更好地进行多任务学习. 可塑性能够帮助智能体动态地调整权重,在没有明确的奖励信号的情况下, 学会训练过程中未见过的场景, 例如转弯等. 
    \caption{\textbf{Diagram of PDLF.} Top: By combining Synaptic Cooperation Plasticity (SCP) and Presynaptic-Dependent Plasticity (PDP), neurons can achieve diverse and heterogeneous plasticity. Bottom: Agents with PDLF learn plasticity rather than directly adjusting weights. Different forms of synaptic plasticity can be formed between neurons, enabling better multi-task learning. Plasticity helps the agents dynamically adjust weights and learn previously unseen scenarios during training, even without explicit reward signals.}\label{abstract}
\end{figure}

% Sample body text. Sample body text. Sample body text. Sample body text. Sample body text. Sample body text. Sample body text. Sample body text.

\subsection{Plasticity-Driven Learning Framework}\label{sec2_sub1}

% 尖峰神经网络 (SNN) 领域见证了多种局部可塑性机制的识别，例如尖峰时序相关可塑性 (STDP) 和 Bienenstock-Cooper-Munro (BCM) 规则。 然而，将这些生物学观察结果抽象并提炼为适合各种复杂任务的具体可塑性机制仍然是一个艰巨的挑战。 为了避免过多的手动设计和不同可塑性机制的简单组合，我们引入了更抽象的可塑性级别：参数化可塑性框架。
The realm of Spiking Neural Networks (SNNs) has witnessed the identification of multiple local plasticity mechanisms, such as Spike-Timing-Dependent Plasticity (STDP)~\cite{bi1998synaptic} and Bienenstock-Cooper-Munro (BCM) rules~\cite{gjorgjieva2011triplet}. However, abstracting and refining these biological observations into concrete plasticity mechanisms suitable for diverse complex tasks remains a formidable challenge. To circumvent excessive manual design and simple combination of different plasticity mechanisms, we introduce a more abstract level of plasticity: a parametric plasticity framework.

% 遵循局部可塑性的基本规则——突触强度受到突触前和突触后神经活动的影响——我们基于 Spiking BCM 模型的参数扩展设计了我们的框架。 以可塑性为中心的学习框架包括两个关键组成部分：突触合作可塑性（SCP）和突触前依赖性可塑性（PDP）。 SCP 通过考虑突触前和突触后神经元的活动来动态调整突触强度。 相比之下，PDP 仅根据突触前活动进行调整，并引入了对突触变化的偏差以实现稳定性。 我们框架中的突触权重更新公式如下：
Adhering to the fundamental rule of local plasticity - that synaptic strength is influenced by pre- and post-synaptic neural activity - we design our framework based on a parametric expansion of the Spiking BCM rule~\cite{bekolay2013simultaneous}. The PDLF comprises two key components: Synaptic Cooperation Plasticity (SCP) and Presynaptic-Dependent Plasticity (PDP). SCP dynamically adjusts synaptic strength by considering the activity of both pre- and post-synaptic neurons. In contrast, PDP adjusts based on pre-synaptic activity alone and introduces a bias to synaptic changes for stability. The synaptic weight update in our framework is formulated as follows:

\begin{equation}
    \Delta w_{i,j} = \eta \ (\underbrace{A_{i,j} x_i (x_j - C_{i,j})}_{\text{Synaptic Cooperation Plasticity (SCP)}} + \underbrace{B_{i,j} x_j + D_{i,j}}_{\text{Presynaptic-Dependent Plasticity (PDP)}})
    \label{plasticity_centric}
\end{equation}

% 在方程~\ref{ Plasticity_centric}中，$\Delta w_{i,j}$表示神经元$i$和$j$之间突触权重的变化。 x_i和x_j 则分别表示突触前后神经元的脉冲迹. 参数$A_{i,j}$、$B_{i,j}$、$C_{i,j}$和$D_{i,j}$是可学习的，使网络能够形成独特且适应性强的可塑性 规则。 SCP，由术语 $A_{i,j} x_i (x_j - C_{i,j})$ 表示，根据神经活动的时间相关性调整突触强度，其中 $C_{i,j}$ 充当 突触后活动的阈值。 PDP，用 $B_{i,j} x_j + D_{i,j}$ 表示，根据突触前活动修改突触权重，$D_{i,j}$ 为每个神经元提供稳定的偏差。 学习率 $\eta$ 缩放整体突触权重变化。
In Eq.~\ref{plasticity_centric}, $\Delta w_{i,j}$ represents the change in synaptic weight between neurons $i$ and $j$. $x_i$ and $x_j$ represent the spike traces~\cite{pfister2006triplets} of pre- and post-synaptic neurons respectively. The parameters $A_{i,j}$, $B_{i,j}$, $C_{i,j}$, and $D_{i,j}$ are learnable, enabling the network to form distinct and adaptable plasticity rules. SCP, represented by the term $A_{i,j} x_i (x_j - C_{i,j})$, adjusts synaptic strength based on the temporal correlation of neural activities, with $C_{i,j}$ acting as a threshold for post-synaptic activity. PDP, denoted by $B_{i,j} x_j + D_{i,j}$, modifies synaptic weights based on pre-synaptic activity, with $D_{i,j}$ providing a stable bias for each neuron. The learning rate $\eta$ scales the overall synaptic weight change.

% 为了优化这些参数，我们采用了进化策略（ES）~\cite{sehnke2010parameter}，其灵感来自塑造生物有机体的自然选择过程。 在这种情况下，以可塑性为中心的学习规则的参数可以被视为内在先验，在整个进化过程中进行优化以确保生存和适应。 ES 涉及一组智能体，每个智能体都有一组独特的可塑性参数，并根据各种任务的适应性和表现来评估它们的适应性。 通过这个过程，参数得到优化，使代理能够在其整个生命周期中保持灵活和动态的适应。这种对于局部可塑性的参数化推广, 这突破了以往单一的可塑性建模, 能够为每个突触分配独立的, 异质性的可塑性规则. 同时, 这将使代理人即使在没有明确奖励信号的情况下也能保持可塑性, 并动态地适应环境.

To optimize these parameters, we employ an Evolutionary Strategy (ES)~\cite{sehnke2010parameter}, inspired by the natural selection processes shaping biological organisms. In this context, the parameters of the plasticity-centric learning rule can be viewed as intrinsic priors, optimized throughout the evolutionary process to ensure survival and adaptation. The ES involves a population of agents, each with a unique set of plasticity parameters, with their fitness evaluated based on adaptability and performance in various tasks. Through this process, the parameters are optimized, enabling agents to maintain flexible and dynamic adaptation throughout their lifespan.

\subsection{PDLF Enhances Working Memory Capacity}\label{sec2_sub2}
% TODO: 添加p检验?

\begin{figure}[htbp]
    \centering
    \includegraphics[width=\textwidth]{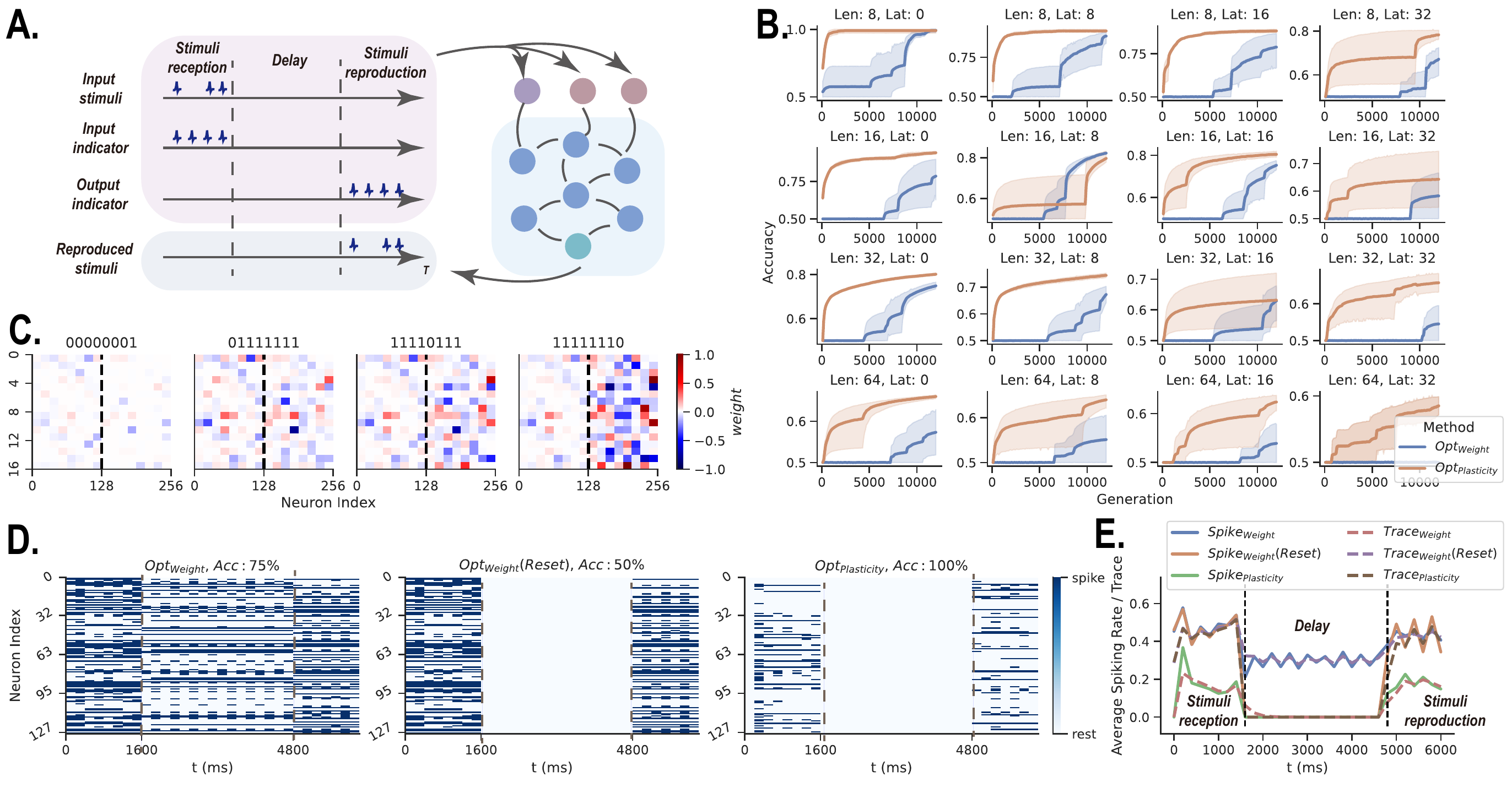}
    % 工作记忆实验的设计以及元可塑性对于工作记忆能力的影响. A. DMS任务的示意图. SNNs首先接受一串运动样本的刺激, 每一个样本持续 200ms, 随后是不同长度的延迟期, 最后是一个和样本刺激持续时间相同的测试刺激. SNNs需要在接收到测试刺激的时候按顺序重复第一阶段受到的刺激. B. 具有可塑性的SNNs和权重直接训练的SNNs的性能比较. 具有可塑性的SNNs有更快的收敛速度, 更长的记忆时间以及更大的记忆容量. Len指的是刺激样本的持续的步, 而Lat则指的是延迟期的步数. C. 在运动样本为$8$时, 不同的运动刺激输入结束后的突触权重. 具有可塑性的SNNs能够针对不同的刺激, 形成不同的连接权重. 虚线的左边是和刺激相关的输入权重, 右边是输出权重. D. 权重直接训练的SNNs和元可塑性的SNNs在不同阶段的神经元状态. 直接训练的SNNs需要延迟期的神经活动来维持记忆, 在输入刺激之后将神经元的膜电位置为$0$会使得记忆的正确率回到机会水平, 失去记忆能力. 而具有可塑性的SNNs则能够将输入刺激编码到突触权重中, 具有更强大的记忆功能. E. 使用不同策略的SNNs在不同阶段的脉冲率以及平均的脉冲迹的可视化. 具有可塑性的SNNs能够保持更低的脉冲率. 
    \caption{Design of the WM experiment and the impact of PDLF on WM. \textbf{A.} Schematic of the copying task. The SNNs first receive a sequence of motion stimuli, with each stimulus lasting $200$ ms, followed by a delay period of varying lengths, and finally, a test stimulus of the same duration as the sample stimulus. The SNNs are required to reproduce the stimuli from the first phase to receive the test stimulus. \textbf{B.} Performance comparison between SNNs with plasticity and trained with direct weights. SNNs with plasticity show faster convergence, longer memory duration, and greater memory capacity. 'Len' refers to the length of stimulus samples, while 'Lat' refers to the number of steps in the delay period. \textbf{C.} Synaptic weights after different motion stimulus inputs when the number of motion samples is $8$. SNNs with plasticity can form distinct connection weights for different stimuli. The left side of the dashed line shows the input weights associated with the stimulus, and the right side shows the output weights. \textbf{D.} Neuron states at different stages in SNNs trained directly with weights and those with PDLF. Directly trained SNNs require neural activity during the delay period to maintain memory. Resetting the membrane potential to $0$ after the input stimulus leads to a chance-level memory accuracy, resulting in memory loss. In contrast, SNNs with plasticity can encode input stimuli into synaptic weights, demonstrating stronger memory functionality. \textbf{E.} Visualization of the firing rates at different stages and the average spike traces for SNNs using different strategies. SNNs with plasticity can maintain lower firing rates.}
    \label{memory}
\end{figure}

% 在本节中，我们旨在证明元可塑性对代理人的工作记忆能力的影响。工作记忆是指我们暂时维持和处理信息的能力，是高级智力的基石. 
In this section, we aim to demonstrate the effect of PDLF on Working Memory (WM). WM is the ability to maintain and process information temporarily and is the cornerstone of higher intelligence~\cite{baddeleyWorkingMemory1974}.

% 在这一节中，我们深入研究了元可塑性在提高尖峰神经网络（SNN）工作记忆能力方面的影响作用。我们利用一个被称为延迟匹配采样（DMS）的任务，如图[图参]A所示。在这个任务中，SNNs首先接受一连串的刺激，每个刺激持续200ms。然后是一个不同长度的延迟期，最后是一个与样本刺激持续时间一致的测试刺激。SNNs的目标是在收到测试刺激后按顺序重现初始阶段的刺激。
%We delve into the influential role of PDLF in enhancing the WM capacity of spiking neural networks (SNNs).
We explore the significant impact of PDLF on enhancing the WM capabilities of SNNs. We utilize a task known as the copying task~\cite{hochreiter1997long}, as illustrated in Fig.~\ref{memory}\textbf{A}.  In this task, SNNs are initially presented with a sequence of stimuli, each lasting for $200$ ms. This is then followed by a delay period of variable lengths, and finally, a test stimuli of equivalent duration to the sample stimuli is presented. The challenge for the SNNs lies in accurately reproducing the initial sequence of stimuli in the correct order upon receiving the test stimuli.

% 为了充分展示基于元弹性的方法在工作记忆任务中的优势，我们在广泛使用的三层SNNs的基础上，与直接利用ES优化权重的策略进行了比较。如图所示，具有元弹性的SNN表现出更快的收敛速度、更长时间的记忆保持能力以及更强的记忆容量。
To thoroughly demonstrate the advantages of employing a PDLF-based approach in working memory tasks, we carry out a comparison with the strategy of directly optimizing weights utilizing the ES based on widely used three-layer SNNs. As shown in Fig.~\ref{memory}\textbf{B}, SNNs equipped with PDLF exhibit faster convergence rates, an ability to retain memory over longer durations, and an enhanced memory capacity.

% 为了进一步研究元可塑性的影响，我们研究了不同运动刺激输入后的突触权重，当运动样本的数量被设置为8美元。如图[图参]C所示，具有可塑性的SNNs可以针对不同的刺激形成不同的连接权重，从而提供了它们优越的适应性的证据。
To further investigate the influence of PDLF, we visualize synaptic weights following various stimuli inputs when the stimuli length is set to $8$. As shown in Fig.~\ref{memory}\textbf{C}, SNNs endow with PDLF can form distinct connection weights for different stimuli, demonstrating their superior adaptive capacity.

% 直接训练的SNNs和具有元可塑性的SNNs在不同阶段的神经元状态也得到了比较（图~ref{memory}\textbf{D}）。直接训练的SNNs在延迟期需要神经活动来维持记忆，在刺激输入之后将膜电位重置为$0$之后, SNNs就失去了对于输入刺激的记忆, 这说明对于直接优化权重的SNNs, 记忆主要存储在神经活动之中. 但具有可塑性的SNNs设法将输入刺激编码为突触权重，从而展示了卓越的记忆功能。除此之外，具有可塑性的SNNs调整突触权重以促进工作记忆的能力使神经元在不接收任务相关刺激时保持静止状态。这有助于提高网络效率，增强网络容量。这种通过可塑性实现高效记忆的方法在其他生物和计算神经科学研究中也得到了验证。然而，我们的工作记忆实验与众不同地比较了具有可塑性的SNNs和直接训练的SNNs在更复杂任务中的表现差异，从而强调了元可塑性的重要性。我们的结果清楚地表明，元可塑性在提高人工神经网络的性能和效率方面发挥着重要作用。
We compare the neuronal states at various stages between SNNs directly trained with weights and those incorporating PDLF, as illustrated in Fig.~\ref{memory}\textbf{D}. SNNs directly trained with weights rely on neuronal activity during the delay period to maintain memory. Resetting the membrane potential to $0$ after the stimulus input leads to memory loss for input stimuli. This indicates that in SNNs directly optimized with weights, memory is primarily stored in neuronal activity. In contrast, SNNs incorporating PDLF encode the input stimulus into synaptic weights, demonstrating remarkable memory capabilities. Their ability to adjust synaptic weights facilitates the enhancement of working memory and allows neurons to remain in a resting state when not receiving task-related stimuli. This contributes to network efficiency and enhances network capacity, which has been validated in other biological and computational neuroscience studies~\cite{mongillo2008synaptic, masseCircuitMechanismsMaintenance2019a}. %However, our working memory experiment distinctively compared the performance differences between SNNs with plasticity and directly trained SNNs in more complex tasks, thereby emphasizing the importance of PDLF. Our results provide a clear demonstration of the significant role that PDLF plays in improving the performance and efficiency of SNNs.

% 最后，我们还检查了不同阶段的发射率，并可视化了采用不同策略的SNNs的平均尖峰轨迹。如图[图参]E所示，具有可塑性的SNNs可以保持较低的发射率，有助于提高其管理计算资源的效率。
Finally, we visualize the average spike traces of SNNs under different training strategies, as shown in Fig.~\ref{memory}\textbf{E}. Notably, SNNs incorporating PDLF can sustain lower firing rates, thereby enhancing their efficiency in managing computational resources.

% 总之，这些发现强调了元可塑性在增强SNNs的工作记忆能力方面所起的实质性作用，显示了它在促进人工智能系统的复杂认知任务方面的潜力。
In summary, these results highlight the significant role of PDLF in bolstering the working memory capacity of SNNs, demonstrating its potential for facilitating complex cognitive tasks in artificial intelligence systems.

% 元可塑性增强了多任务学习能力
% \subsection{PDLF in Multi-task Reinforcement Learning}\label{sec2_sub3}
\subsection{PDLF Enhances Multi-task Learning}\label{sec2_sub3}

% 强化学习（RL）涉及一个代理学习与环境互动以实现某个目标，其行动的质量由奖励信号决定。传统的强化学习方法通常需要在一个固定奖励函数的单一任务上训练代理。然而，在复杂的现实世界环境中，代理人需要处理多个任务，并适应不断变化的奖励函数。这种多任务学习的情况带来了巨大的挑战，特别是当所涉及的任务是多样化和潜在的冲突。
Reinforcement Learning (RL) involves an agent learning to interact with its environment to achieve a specific goal, with the quality of its actions dictated by a reward signal. Conventional RL approaches often entail training the agent on a single task with a fixed reward function. However, in complex, real-world environments, agents need to handle multiple tasks and adjust to changing reward functions. This multi-task learning scenario presents significant challenges, mainly when the tasks involved are diverse and potentially conflicting.

\begin{figure}[t]
    \centering
    \includegraphics[width=\textwidth]{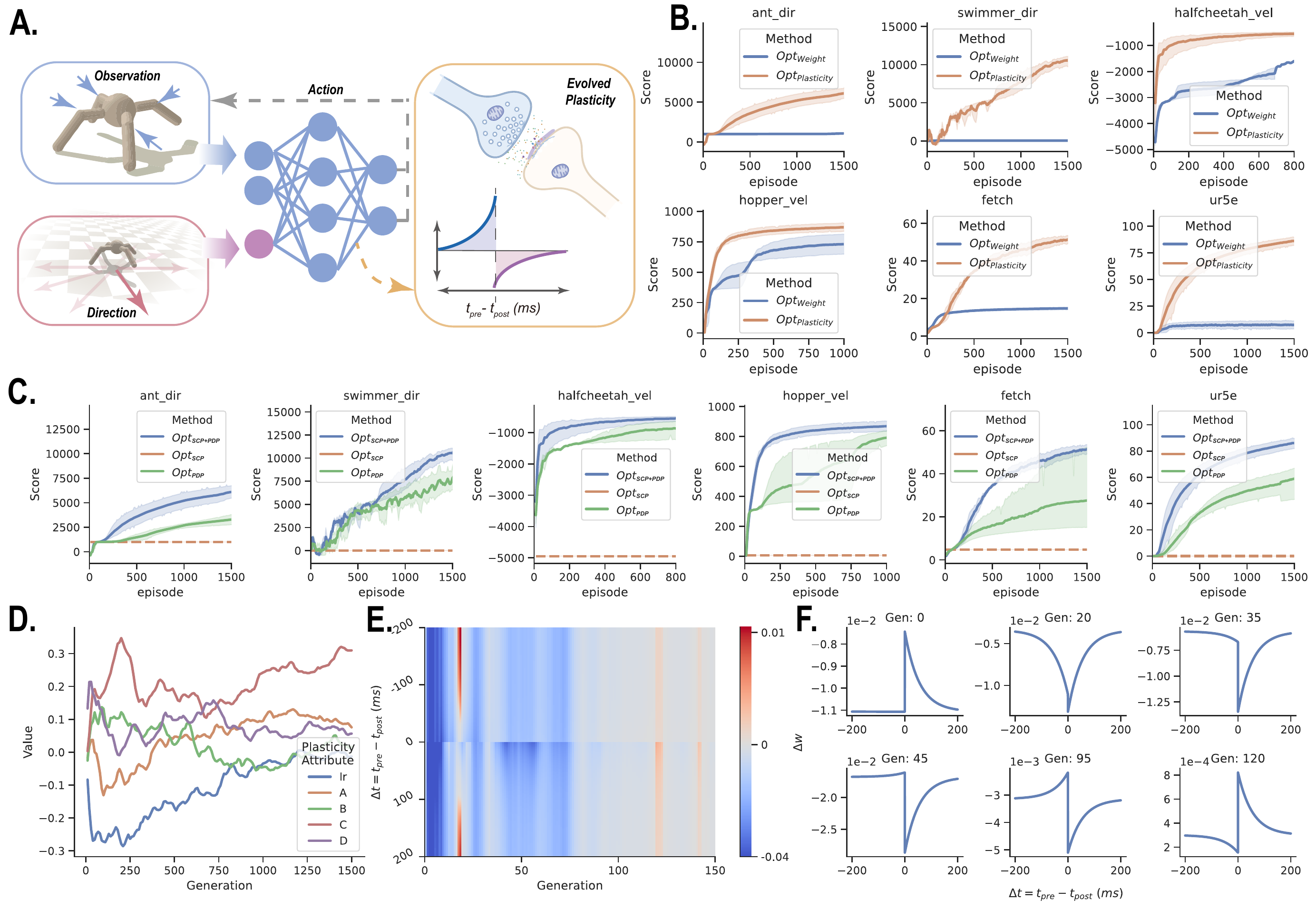}
    % 元可塑性在多任务强化学习任务中的表现以及不同的可塑性属性, 神经元损伤对于性能的影响. A. 多任务强化学习的示意图. 智能体需要使用单一的网络同时学会目标不同, 甚至完全相反的多个任务. 任务的目标被视作智能体的观测, 并和其他的观测(例如关节的位置, 速度等)同时被输入到SNNs. A. 具有元可塑性的智能体和权重直接训练的智能体的训练曲线. 在这些多任务强化学习任务中, 智能体需要学会以不同的方向(ant_dir, swimmer_dir), 不同的速度(halfcheetah_vel, hopper_vel)以及不同的位置(fetch, ur5e)为目标运动. 具有元可塑性的智能体能在多任务挑战中, 保持动态的突触权重, 学习不同任务的特点, 进而实现更好的性能. B. 不同的智能体在训练任务和训练过程中未见过的任务中的表现. 具有可塑性的智能体能够很好地泛化到训练过程中未见过的任务, 而权重直接训练的智能体由于在测试过程中, 权重固定, 难以泛化到训练中未见过的测试任务. C. 不同的可塑性机制的消融分析. 不同颜色代表了删除了部分可塑性(STDP, Synaptic Redistribution 或者 Synaptic Scaling)的训练曲线. 在删除了Synaptic Redistribution后, 由于SNNs失去了稳态机制, 导致SNNs发散. 而STDP以及Synaptic Scaling则都对于提高智能体的性能具有重要的作用. D. 在面对临时的神经损伤时候, 智能体的表现. 在第500个step时候, 所有的突触权重为被重置为$0$用于模拟神经系统突然遭受损伤, 并持续$50$个step. 具有可塑性的智能体能够从这种临时的损失中恢复, 具有更好的鲁棒性. E. 在面对不同程度的永久的神经损伤时, 智能体的表现. 在测试的初始阶段, 不同比例的神经元被屏蔽, 其权重被设置为$0$, 并无法更新, 用于模拟神经网络的永久损伤. 具有可塑性的智能体能够在应对这种永久性损伤时候, 具有更高的性能和鲁棒性. F. 对于智能体泛化性能的测试. 在只改变智能体的目标方向时, 智能体的运动轨迹. 绿色是智能体匀速运动的期望轨迹. 蓝色是具有可塑性的智能体的实际运动轨迹. 具有可塑性的智能体能够更好地理解不同的任务, 并能够应对不同任务复合的复杂场景, 具有更好的泛化能力. G. 一个突触的元可塑性在训练过程中的变化曲线, 通过演化策略, 智能体学会了调整自身的可塑性. H. 在不同的突触前后神经元的脉冲时间间隔时, 训练过程中, 不同代智能体的可塑性对于权重的影响. I. H中不同代智能体中的具体的可塑性的功能. 
    \caption{\textbf{PDLF's performance in multi-task reinforcement learning tasks and the influence of different plasticity attributes on performance.} \textbf{A.} Illustration of multi-task reinforcement learning. The agent is required to utilize a singular network to simultaneously learn multiple tasks with distinct objectives or even entirely opposing ones. The objectives of the tasks are treated as observations for the agent. They are inputted to the SNNs along with other observations such as joint positions, velocities, etc. \textbf{B.} Training curves of agents with PDLF versus those trained directly on weights. In these multi-task RL tasks, agents need to learn to move towards different directions (\texttt{ant\_dir}, \texttt{swimmer\_dir}), at varying speeds (\texttt{halfcheetah\_vel}, \texttt{hopper\_vel}), and to different locations (\texttt{fetch}, \texttt{ur5e}). Agents with PDLF maintain dynamic synaptic weights, learn characteristics of different tasks, and hence achieve superior performance in multi-task challenges. \textbf{C.} Ablation analysis of different plasticity mechanisms. Different colors represent training curves with some form of plasticity (SCP or PDP) removed. After removing PDP, SNNs diverge due to the loss of the equilibrium mechanism. Both SCP and PDP play significant roles in enhancing agent performance. \textbf{D.} The change curve of a synapse's PDLF during the training process. Through evolutionary strategies, agents learn to adjust their plasticity. \textbf{E.} During the training process, at different inter-spike intervals of pre-synaptic and post-synaptic neurons, the impact of the plasticity of agents from different generations on weights. \textbf{F.} The specific functions of plasticity in agents from different generations as shown in \textbf{E}.}\label{rl_exp}
\end{figure}

% 在RL领域，一些最具挑战性的问题位于连续控制领域，通常使用复杂的物理引擎（如MuJoCo或Brax）进行建模。这些任务要求代理以高精确度和协调性操纵模拟的物理实体，类似于人类控制他们的四肢来执行复杂的任务。我们利用Brax模拟器来设计六个不同的连续控制环境。在这些环境中，代理人需要以各种速度、方向和目的地点进行导航。不同的任务目标被作为观测以指导智能体完成不同的任务. 这种具有挑战性的任务被作为元学习等领域的基线. 在训练过程中，他们只接触到有限的任务实例，如八个特定的方向或八个固定的速度，并使用单一的网络来学习这些不相关的，甚至是相互冲突的任务。
In the field of RL, some of the most challenging problems lie within the domain of continuous control, usually modeled using sophisticated physics engines. These tasks require agents to manipulate simulated physical entities with high precision and coordination, similar to how humans control their limbs to carry out complex tasks. We use the Brax~\cite{brax2021github} simulator to design six continuous control environments. In these settings, agents need to navigate at various speeds, directions, and destination points. As shown in Fig.~\ref{rl_exp}\textbf{A}, different task objectives are treated as observations to guide the agent in accomplishing different tasks. These challenging tasks serve as baselines in fields such as meta-learning~\cite{finn2017model, zhou2022domain, ibarz2021train}. During training, they are only exposed to a limited number of task instances, such as eight specific directions or eight fixed speeds. They use a single network to learn these unrelated or conflicting tasks.

% 我们的实验涉及两种类型的尖峰神经网络（SNNs）之间的比较--一种是直接优化的突触权重，另一种是优化的可塑度。这两种类型的SNNs都保持相同的规模和结构，以确保公平的比较。通过这一点，我们旨在阐明在多任务环境下，优化可塑性比直接优化权重的优势。这种比较也有助于评估我们提出的模型的有效性和潜力，特别是在面对复杂的连续控制任务所带来的固有挑战时。
Our experiments compare two types of SNNs - one with synaptic weights that have been directly optimized and another with optimized PDLF. Both types of SNNs maintain the same scale and structure to ensure a fair comparison. Through this, we aim to highlight the advantages of optimizing PDLF over direct weight optimization in a multi-task environment. This comparison also evaluates the effectiveness and potential of our proposed model, especially in the face of the inherent challenges posed by complex continuous control tasks.

\begin{table}[h]
    % 比较各种强化学习任务的性能，在一个具有128个隐藏尖峰神经元的三层全连接SNN中，突触可塑性的不同配置。每个任务都使用不同的训练方法进行评估： 可塑性$_{ABCD}$、可塑性$_{ACD}$、可塑性$_{BD}$、可塑性$_{ABC}$和直接重量训练。显示的数值是多次试验的平均值和标准差。各项任务的性能指标没有直接的可比性。机会水平 "一行显示了随机选择行动时的性能指标。值得注意的是，在去除突触再分布可塑性后（如在可塑性$_{ABC}$条件下），SNNs变得不稳定，导致分歧。
    \caption{Comparison of performance across various reinforcement learning tasks with different configurations of synaptic plasticity. Each task is evaluated using different training methods: Plasticity$_{SCP+PDP}$, Plasticity$_{SCP}$, Plasticity$_{PDP}$, and direct weight training. The values presented are the mean and standard deviation over $5$ trials. The row 'Chance Level' shows the performance metrics when randomly chooses actions. $^*$ Upon the removal of Presynaptic-Dependent Plasticity (PDP), the SNNs become unstable, leading to divergence.}\label{tab:rl_exp}%
    \resizebox{\linewidth}{!}{
        \begin{tabular}{ccccccc}
            \toprule
            Training                 & \texttt{ant\_dir} & \texttt{swimmer\_dir} & \texttt{halfcheetah\_vel} & \texttt{hopper\_vel} & \texttt{fetch} & \texttt{ur5e} \\
            \midrule
            \textbf{Opt$_{SCP+PDP}$} & $6904 \pm 801$    & $10531 \pm 827$       & $-549 \pm 95$             & $869 \pm 38$         & $51 \pm 3$     & $86 \pm 5$    \\
            % P$_{ACD}$                 & $4464 \pm 544$ & $6906 \pm 1449$ & $-550 \pm 62$    & $670 \pm 250$ & $28 \pm 21$  & $38 \pm 7$  \\
            Opt$_{PDP}$              & $3284 \pm 570$    & $7831 \pm 1479$       & $-870 \pm 312$            & $792 \pm 50$         & $26 \pm 26$    & $58 \pm 13$   \\
            Opt$_{SCP}^{*}$          & -                 & -                     & -                         & -                    & -              & -             \\
            Opt$_{Weight}$           & $1069 \pm 98$     & $31 \pm 8$            & $-1598 \pm 324$           & $729 \pm 116$        & $15 \pm 0.6$   & $7 \pm 5$     \\
            \midrule
            Chance Level             & $995 \pm 0.01$    & $0.12 \pm 0.02$       & $-4946 \pm 1.3$           & $6.51 \pm 0.3$       & $4.74 \pm 0.0$ & $0 \pm 0.0$   \\
            % row 3    & data 7     & data 8                 & data 9\footnotemark[2]                              \\
            \bottomrule
        \end{tabular}
    }
    % 在移除突触重分布可塑性之后, SNNs变得不稳定, 会导致发散.
    % \footnotetext{Source: This is an example of table footnote. This is an example of table footnote.}
    % \footnotetext[2]{Example for a second table footnote. This is an example of table footnote.}
\end{table}

% 我们在具有128个隐层脉冲神经元的三层的全连接SNN模型上研究了元可塑性的表现, 这是一个广泛使用的结构. 具体的方法就是使用演化策略优化可塑性和突触权重, 并比较这两种策略在多任务连续机器人控制任务上的表现. 突触权重以及可塑性的参数都被初始化为$0$. 在测试过程中, 对于具有可塑性的智能体, 其可塑性规则将被固定, 突触权重会被重置为$0$, 而对于直接训练权重的智能体, 其训练权重将被应用于测试. 我们使用了不同的强化学习任务以全面地对元可塑性进行测试, 这些任务要求智能体同时学会应对不同的任务, 并将学到的知识泛化到未见过的, 更复杂的任务上. 
We explore the performance of PDLF in a three-layer, fully-connected SNN model with $128$ hidden spiking neurons. %Specifically, we optimized the plasticity and synaptic weights using ES~\cite{sehnke2010parameter} and compared the performance of these two strategies on multi-task continuous robotic control tasks. 
Both the synaptic weights and the plasticity parameters are initialized to $0$. During testing, their plasticity rules are fixed for agents with plasticity, and synaptic weights are reset to $0$. The trained weights are applied during testing for agents trained directly on weights. We utilize reinforcement learning tasks to thoroughly test PDLF, requiring the agents to learn to cope with different tasks simultaneously and generalize the acquired knowledge to unseen, more complex tasks.

% 在我们的实验过程中，与直接优化突触权重的代理相比，具有元可塑性的代理在多个任务中表现出更优越的性能。如图~ref{rl_exp}A所示，元可塑性代理表现出更有效的学习曲线。元弹性代理迅速适应了任务的变化，使它们更有能力应对设计环境中固有的多任务挑战。相比之下, 直接训练权重的SNNs则无法适应不同的任务, 只能学习到一些平凡的解, 例如在多个目标方向的任务中, 保持近似静止.
In our experiments, agents with PDLF show superior performance in multiple tasks compared to agents with directly optimized synaptic weights. As seen in Fig.~\ref{rl_exp}\textbf{B} and Tab.~\ref{tab:rl_exp}, agents with PDLF exhibit more effective learning curves. The agents with PDLF quickly adapted to the changes in tasks, making them more capable of tackling the multi-task challenges inherent in the designed environments. In contrast, SNNs with directly trained weights fail to adapt to different tasks and can only acquire trivial solutions, such as maintaining approximate immobility in tasks involving multiple target directions.

% 图~ref{rl_exp}C中描述的消融研究提供了对不同可塑性机制贡献的进一步了解。去除任何形式的可塑性都会导致代理人的表现下降，而去除HP则会因为平衡机制的丧失而导致分歧。这一结果强调了所有三种可塑性机制，即STDP、突触缩放和突触再分配，在维持代理人的稳定性和适应性方面的重要性。
Ablation studies, depicted in Fig.~\ref{rl_exp}\textbf{C} and Tab.~\ref{tab:rl_exp}, provide further insights into the contributions of different plasticity mechanisms. Removing any form of plasticity results in decreased agent performance, with the removal of PDP causing a divergence due to the loss of the equilibrium mechanism. This result highlights the importance of all plasticity mechanisms in maintaining the stability and adaptability of the agents.

% 在训练过程中，突触的元可塑性的变化曲线（图~ref{generation}G），以及可塑性对不同代的代理权的影响（图~ref{generation}H），表明在进化策略的推动下，有效地学习了元可塑性规则的最佳参数。有趣的是，不同代的代理的可塑性的具体功能是不同的（图~ref{generation}I），表明可塑性机制的进化和微调，以提高各代的代理性能。
The changing curve of a synapse's PDLF during training (Fig.~\ref{rl_exp}\textbf{D}), along with the impact of plasticity on weights for different generations of agents (Fig.~\ref{rl_exp}\textbf{E}), demonstrate the effective learning of optimal parameters for the PDLF rule, facilitated by the evolutionary strategy. Interestingly, the specific functions of plasticity across different generations of agents differed (Fig.~\ref{rl_exp}\textbf{F}), indicating the evolution and fine-tuning of plasticity mechanisms to improve agent performance across generations.

% 元可塑性增强泛化能力
\subsection{PDLF Enhances Generalization Ability}\label{sec2_sub4}

\begin{figure}[htbp]
    \centering
    \includegraphics[width=\textwidth]{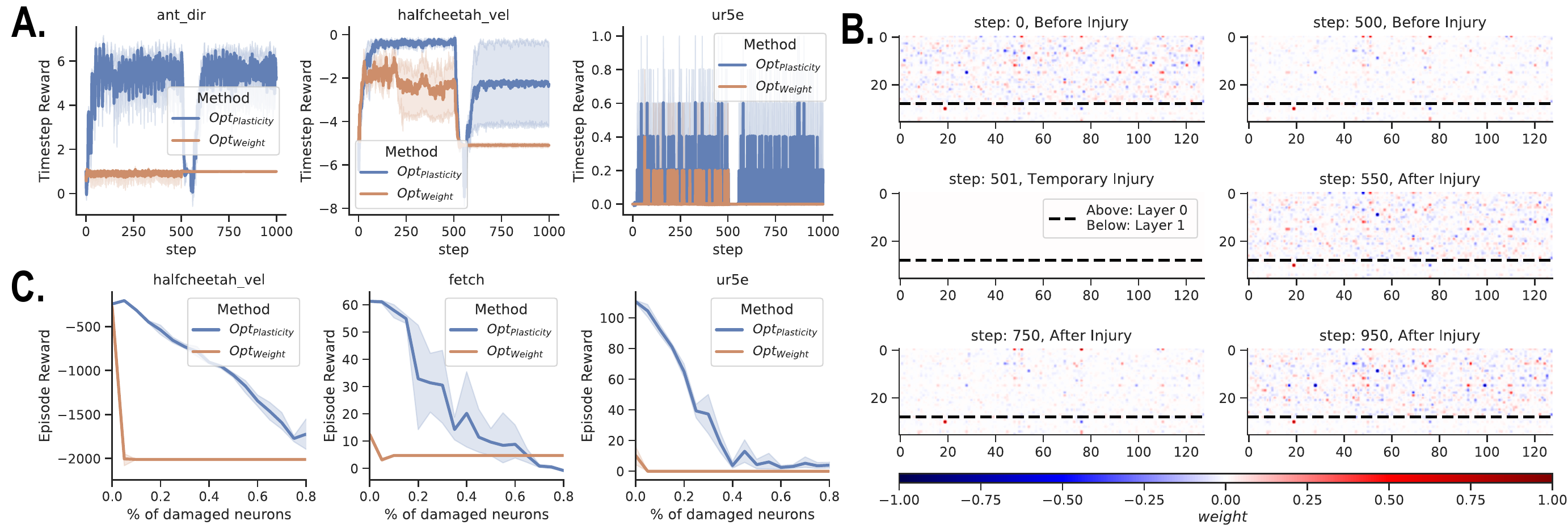}
    % 元可塑性能够提升智能体的泛化能力, 使得智能体在面对未免过的任务, 以及神经元损伤时, 具有更强的性能. A. 在面对临时的神经损伤时候, 智能体的表现. 在第500个step时候, 所有的突触权重为被重置为$0$用于模拟神经系统突然遭受损伤, 并持续$50$个step. 具有可塑性的智能体能够从这种临时的损失中恢复, 具有更好的鲁棒性. B. 在临时的损伤前后, 不同的时刻网络的权重. 虚线的上面是输入层的突触权重, 而虚线的下面是读出层的权重. 即使智能体在遭遇临时的损伤丢失了全部的突触权重, 智能体仍能够根据可塑性以及输入刺激, 恢复突触权重. C. 在面对不同程度的永久的神经损伤时, 智能体的表现. 在测试的初始阶段, 不同比例的神经元被屏蔽, 其权重被设置为$0$, 并无法更新, 用于模拟神经网络的永久损伤. 具有可塑性的智能体能够在应对这种永久性损伤时候, 具有更高的性能和鲁棒性.
    \caption{\textbf{Performance under temporary and permanent nerve injury.} \textbf{A.} The agent's performance in the face of temporary neural damage. At the $500$th step, all synaptic weights were reset to $0$ to simulate a sudden neural system injury, and this condition lasted for $50$ steps. Agents with plasticity were able to recover from this temporary loss, demonstrating better robustness. \textbf{B.} Network weights at different times before and after temporary damage. The synaptic weights of the input layer are shown above the dashed line, while the readout layer weights are below the dashed line. Even if the agent loses all synaptic weights due to temporary damage, it can still recover these weights based on its plasticity and input stimuli. \textbf{C.} The agent's performance in the face of permanent neuronal damage of varying degrees. At the start of the test, neurons were blocked at different proportions, their synaptic weights set to $0$, and could not be updated, simulating permanent neural network damage. Agents with plasticity performed better and exhibited stronger robustness when dealing with such permanent damage.}\label{gen0}
\end{figure}

% 元可塑性是增强代理人泛化能力的重要机制，使代理人在处理不熟悉的任务或面临神经元损伤时能表现出更强的性能（图~ref{generation}）。
PDLF serves as an important mechanism for enhancing an agent's generalization abilities, enabling the agent to exhibit stronger performance when dealing with unfamiliar tasks or when facing neuronal damage.

% 我们进一步探究元可塑性在智能体面临损伤时的表现. 我们设计了两种不同的损伤类型: 临时损伤和永久性的损伤. 临时损伤指的是智能体所有的突触权重被重置为$0$, 并持续$50$个步. 永久性损伤指的是部分突触在初始的时候被置为$0$, 并且不会根据可塑性进行更新. 在对损伤的测试中，具有可塑性的代理显示了显著的能力，能够从模拟的临时神经元损伤中恢复，即把所有的突触权重重新设置为0$，持续50步（图~ref{generation}D）。图~ref{generation}B说明了临时损伤前后不同阶段的网络权重变化。值得注意的是，尽管由于造成的临时损伤而失去了所有的突触权重，但代理人设法利用其固有的可塑性和传入的输入刺激恢复这些权重。同时, 即使在永久性神经元损伤的情况下，可塑性强的代理仍然表现出更好的性能和稳健性（图~ref{generation}E）。这些结果表明，元可塑性可以促进人工制剂的复原力，就像它在生物系统中的作用一样。
We further investigate the performance of PDLF when the agents faced injuries. We design two different types of injuries: temporary injuries and permanent injuries. Temporary injuries refer to a scenario where all synaptic weights of the agents are reset to $0$ and kept for $50$ steps. Permanent injuries refer to a situation where some synapses are set to $0$ initially and do not update according to plasticity. In the tests for injuries, agents with plasticity display a remarkable ability to recover from temporary neuronal damage simulated by resetting all synaptic weights to $0$ for $50$ steps (Fig.~\ref{gen0}\textbf{A}). Fig.~\ref{gen0}\textbf{B} illustrates the changes in network weights at various stages before and after the temporary damage. Remarkably, despite losing all synaptic weights due to the inflicted temporary damage, agents manage to recover these weights using their inherent plasticity and incoming input stimuli. Moreover, even under permanent neuronal damage, with a proportion of neurons blocked and their weights unable to update, the plasticity-enabled agents continue to exhibit better performance and robustness (Fig.~\ref{gen0}\textbf{C}). These results suggest that PDLF can contribute to the resilience of artificial agents, much as it does in biological systems.

\begin{figure}[htbp]
    \centering
    \includegraphics[width=\textwidth]{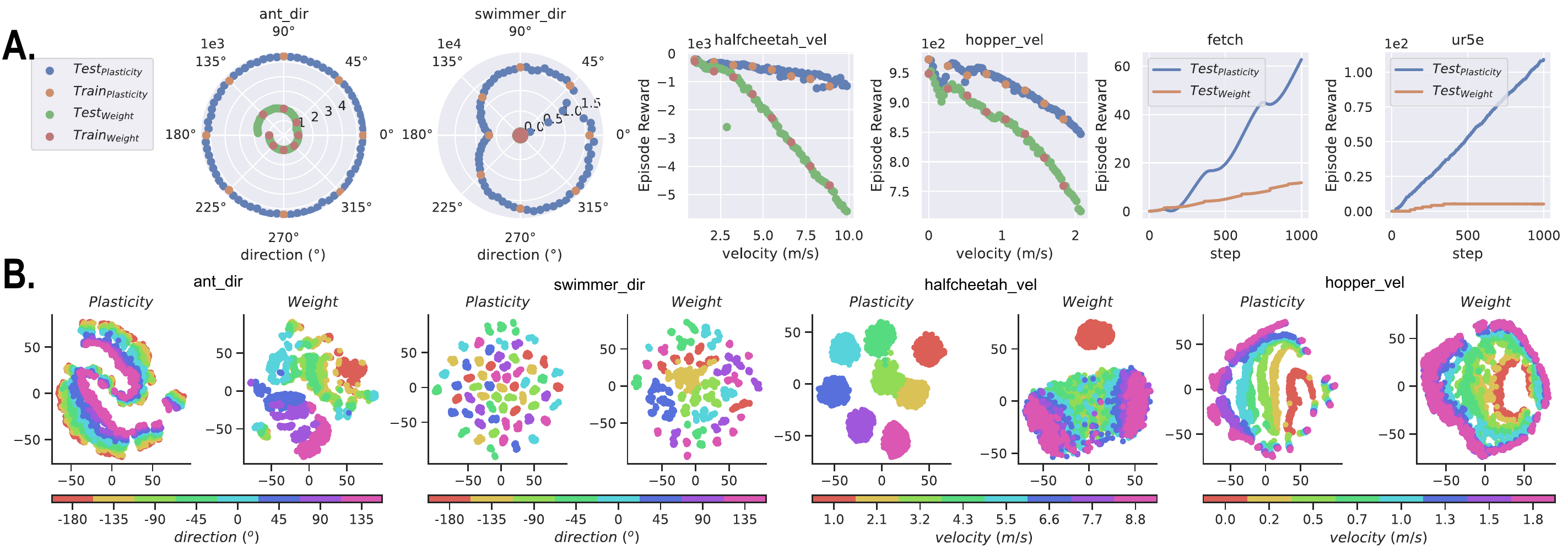}
    % D. 对于智能体泛化性能的测试. 智能体在训练过程中只学会了走直线. 在测试过程中改变智能体的目标方向时, 智能体的运动轨迹. 绿色是智能体匀速运动的期望轨迹. 橙色是具有可塑性的智能体的实际运动轨迹. 具有可塑性的智能体能够更好地理解不同的任务, 针对不同的目标方向, 调整突触的权重, 更加灵活而且具有更好的泛化性能. F. 在强化学习任务上, 使用不同策略训练的, 神经元状态的低维度表示. 每个点代表了一步中的隐藏层神经元的状态. 不同的颜色则代表了不同的任务. 具有可塑性的智能体能够更加区分不同的任务, 相同任务的神经元状态能够形成高维空间上的流形. 
    \caption{\textbf{A.} Performance of different agents in trained tasks and tasks not seen during training. Agents with plasticity can generalize well to unseen tasks, while agents trained directly on weights have difficulty generalizing to unseen test tasks due to their weights being fixed during testing. \textbf{B.} Low-dimensional embeddings of neuronal states during reinforcement learning tasks, differentiated by training strategy. Each point corresponds to the state of the hidden layer neurons at a specific time step. The color coding signifies distinct tasks. Agents that possess plasticity demonstrate an enhanced capability to distinguish between different tasks. Moreover, the neuronal states associated with identical tasks exhibit the intriguing property of forming a manifold within the high-dimensional space.}\label{gen1}
\end{figure}

% 更为突出的是，元弹性代理显示出强大的能力，可以概括训练期间未见过的任务，如图~ref{rl_exp}B所示。对于不同的运动方向以及不同的速度的的任务, 智能体在训练过程中只见过很少的情况, 如图~ref{rl_exp}B中的红色点和橙色的点所示, 因此在这一实验中, 智能体被要求以训练中未见过的方向和速度运动, 这强调了智能体对于任务更深刻的理解和泛化能力. 相比之下，直接根据权重进行训练的代理在测试期间由于其固定的权重而在泛化方面遇到困难。这一观察强调了元弹性所提供的灵活的适应性，增强了代理人在未见场景中的导航能力。
More strikingly, agents with PDLF show a robust ability to generalize to tasks unseen during training, as demonstrated in Fig.~\ref{gen1}\textbf{A}. For tasks with various movement directions and speeds, the agents only encounter a small subset of cases during training, represented by the red and orange points in Fig.~\ref{gen1}\textbf{A}. Therefore, in this experiment, the agents are required to move in directions and at speeds unseen in training, emphasizing the agents' more profound understanding and generalization capacity for the tasks. In contrast, agents trained directly on weights struggle with generalization due to their fixed weights during testing. This observation underscores the flexible adaptability offered by PDLF, enhancing the agent's ability to navigate unseen scenarios.

% 在训练阶段，我们的代理被指示向八个特定方向直线移动。然而，如图xxx所示，具有元弹性的代理人表现出一定程度的概括能力。他们可以学会向训练过程中没有遇到的方向直线移动，而没有元弹性的代理则很难概括他们所学到的东西。
% 为了进一步测试元弹性带来的泛化能力，我们希望代理人能够学会转弯，甚至形成更复杂的路径，仅仅通过改变目标信号，而不需要任何与姿势有关的额外反馈信息。结果显示在图~ref{generation}/textbf{D}。
% 与直接训练其权重的代理相比，那些具有元弹性的代理对这个复杂的任务表现出令人印象深刻的概括能力。它们能够通过元弹性快速调整突触权重，并在不同的训练中动态地修改它们在以前未见过的情况下的状态。这使他们能够朝着不同的目标方向发展。
During the training phase, the agents are instructed to move in straight lines in eight specific directions. However, as illustrated in Fig.~\ref{gen1}\textbf{A}, agents with PDLF demonstrate a degree of generalization capability. They can learn to move in straight lines toward directions not encountered during training, while agents without PDLF struggle to generalize what they have learned.

\begin{figure}[htbp]
    \centering
    \includegraphics[width=\textwidth]{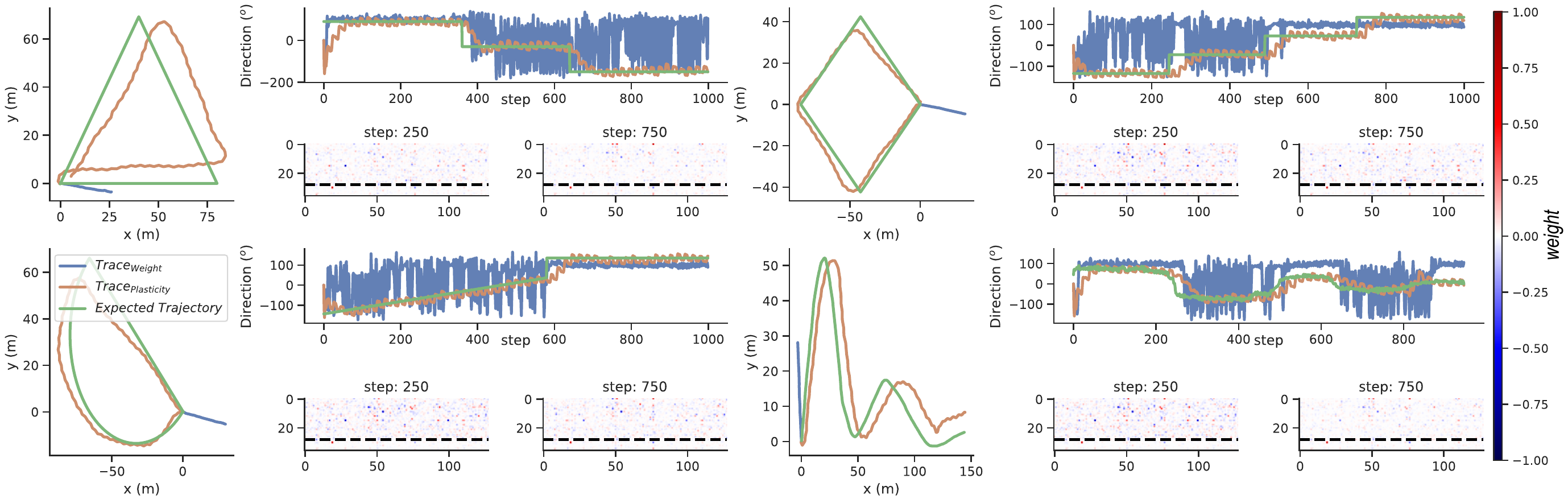}
    % 测试代理的泛化能力。在训练过程中，机器人只学习直线运动。在测试过程中，当目标方向改变时，代理的移动轨迹如图所示。绿线代表以恒定速度移动的预期轨迹。橙色线是具有可塑性的代理的实际运动轨迹。具有可塑性的机器人可以更好地理解不同的任务，根据不同的目标方向调整突触权重，从而表现出更大的灵活性和更优越的泛化性能。
    % TODO: Check
    \caption{\textbf{Testing of the agent's generalization capabilities.} During training, the agent only learned to move in a straight line. The agent's movement trajectories are shown when the target direction is altered during the testing process. The green line represents the expected trajectory of the agent moving at a constant speed. The orange line is the actual movement trajectory of the agent with plasticity. Agents with plasticity can better understand different tasks, adjust synaptic weights according to different target directions, and thus show greater flexibility and superior generalization performance.}\label{gen2}
\end{figure}

To further test the generalization capacity introduced by PDLF, we hope that agents could learn to turn or even form more complex paths simply by changing the target signal and without any additional feedback information related to posture. The results are shown in Fig.~\ref{gen2}.
Compared to agents that directly train their weights, those with PDLF demonstrate impressive generalization abilities toward this complex task. They can quickly adjust synaptic weights through PDLF and dynamically modify their state in previously unseen scenarios during different pieces of training. This allows them to progress toward varying target directions.

% 在图中，我们将低维空间中强化学习任务中使用不同策略训练的代理的神经元状态可视化。图中的每一点都对应于隐藏层神经元在特定时间步的状态，不同的颜色表示不同的任务。这些可视化的显著特点是，具有内在可塑性的代理如何表现出明显的分辨不同任务的能力。更有趣的是，我们观察到与相同任务相对应的神经元状态往往聚集在一起，在高维空间中形成一个清晰的流形。元可塑性的这一特征有助于神经元在保持神经元状态的独特任务特异性模式的同时，在各种任务中具有强大的泛化能力。这进一步说明了具有元可塑性的代理的强大能力，以及这种可塑性对代理的学习和适应能力的深远影响。
In Fig.~\ref{gen1}\textbf{B}, we visualize the neuronal states of agents trained using different strategies during RL tasks in low-dimensional space. Each point in this representation corresponds to the state of the hidden layer neurons at a particular time step, with the varying colors indicating different tasks. The remarkable aspect of these visualizations is how agents with inherent plasticity demonstrate a pronounced ability to discern between distinct tasks. Even more intriguing is the observation that the neuronal states corresponding to the same task tend to cluster together, forming a clear manifold in the high-dimensional space. This feature of PDLF contributes to the agent's robust ability to generalize across various tasks while maintaining distinctive task-specific patterns in neuronal states. This further illustrates the powerful capabilities of agents with PDLF and the profound impact of such plasticity on the agent's learning and adaptation abilities.

\section{Discussion}\label{sec3}
Current artificial intelligence algorithms often focus on solving specific tasks, and their performance may fall short when faced with scenarios that deviate from their training conditions. These algorithms' singular functionality, lack of robustness, and limited flexibility restrict their adaptability and application in complex and variable environments. In contrast, biological entities demonstrate exceptional adaptability in complex and changing environments, typically attributed to the plasticity of biological neural networks. Synaptic plasticity is the cornerstone and heart of exploring more generalized intelligence~\cite{liu2012optogenetic, martin2000synaptic}.

SNNs with their brain-inspired operating mechanisms, lay the foundation for constructing flexible and robust intelligent systems, thereby attracting considerable attention in the field of artificial intelligence~\cite{merollaMillionSpikingneuronIntegrated2014, daviesLoihiNeuromorphicManycore2018, peiArtificialGeneralIntelligence2019, bouvier2019spiking}. However, current SNN training algorithms primarily rely on the backpropagation of external error signals and biologically-inspired plasticity rules, such as Spike-Timing-Dependent Plasticity. Although these methods demonstrate robust performance on individual tasks, the fixed learning paradigm still limits the generalization ability of SNNs and adaptability in multi-task environments. Contrary to the static nature of synaptic weight adjustments in traditional SNNs, we delve into PDLF, facilitated by the synergy between SCP and PDP. PDLF represents a higher-order learning process that dynamically adjusts plasticity rules. Through fostering adaptive synaptic modifications derived from the history of neuronal activity, PDLF promotes the emergence of a more dynamic and self-regulated learning system. Such a mechanism potentially narrows the gap between SNNs and their biological equivalents, thereby facilitating the progression towards continual learning and adaptation.

Our experimental results reveal that PDLF significantly enhances the memory capacity of SNNs by encoding memories directly into synaptic weights. Moreover, it does not rely on spike activity to sustain memory, allowing the network to remain in a resting state when not processing task-related stimuli, thus significantly improving the energy efficiency of SNNs. PDLF also dramatically amplifies the multi-task learning and generalization capabilities of SNNs, facilitating a swift transfer of knowledge learned from other tasks to more complex and unfamiliar tasks.
Regarding adapting the paths and turning towards new directions, our PDLF models can maneuver in ways not encountered during training, a critical feature in complex, dynamic, and unpredictable real-world environments. Importantly, PDLF plays a pivotal role in bolstering the robustness of SNNs under simulated motor impairment scenarios. The exceptional resilience demonstrated in temporary damage scenarios validates the advantages of integrating PDLF into neural networks. Moreover, even in permanent damage, the exhibited resilience reinforces the case for PDLF as an inherent attribute of artificial systems. These characteristics reflect recovery mechanisms in biological systems where the brain mitigates damage through neural reorganization and the formation of new connections.

In conclusion, our study highlights PDLF as a critical feature that enhances the resilience and adaptability of artificial agents. These findings provide valuable insights for designing future artificial systems, opening up new possibilities for creating adaptive, robust, and intelligent agents capable of navigating complex and dynamic environments. Further work can explore more sophisticated forms of PDLF and study their impacts on various facets of artificial agent performance.

\section{Method}\label{sec4}

\subsection{Neuron and synaptic models}

% 在我们的网络模型中，我们采用了漏整合-发射（LIF）神经元，这是因为它们在生物学上的合理性和计算效率。每个LIF神经元的状态由其膜电位表示，它整合输入信号并在电位超过预定阈值时产生尖峰。
We employed leaky integrate-and-fire (LIF) neurons in our network models due to their biological plausibility and computational efficiency. The state of each LIF neuron was represented by its membrane potential, which integrated the incoming signals and generated a spike when the potential crossed a predefined threshold, as shown in Eq.~\ref{lif}.

\begin{equation}
    \tau_m \frac{\partial v}{\partial t} = -(v - v_{0}) + I(t)
    \label{lif}
\end{equation}
% 在这个等式中，$\tau_m$是膜时间常数，$v$是膜电位，$v_{0}$是静息电位，$I(t)$是t$时间的总突触电流。一旦膜电位$v$超过一定的阈值$v_{th}$，神经元就会产生一个尖峰，电位复位到静息值。对神经元行为的离散形式的描述是: 
In Eq.~\ref{lif}, $\tau_m$ is the membrane time constant, $v$ is the membrane potential, $v_{0}$ is the resting potential, and $I(t)$ is the total synaptic current at time $t$. Once the membrane potential $v$ exceeds a certain threshold $v_{th}$, the neuron generates a spike, and the potential is reset.
A discrete form of the LIF neuron's behavior can be described as:

\begin{equation}
    \begin{aligned}
        u(t) & = v(t - \Delta t) + \frac{\Delta t}{\tau_m}(\sum_i w_i s_i(t) - v(t - \Delta t) + v_{0}) \\
        s(t) & = g(u(t) - v_{th})                                                                       \\
        v(t) & =  u(t) (1 - s(t)) + v_{reset} s(t)                                                      \\
        \label{lif2}
    \end{aligned}
\end{equation}

% 在公式~\ref{lif}中，$\Delta t$是时间步长，$v_{reset}$是复位电位，$u(t)$和$v(t)$代表尖峰前和尖峰后的膜电位，$g(\cdot)$是模拟尖峰行为的Heaviside函数。$w_i$ 是突触的权重, 往往被作为优化的参数. 在失去监督信号之后, 对网络权重的更新就会停止. 这种直接优化权重的策略在我们的实验中被设置为对照组. 表2给出了神经元参数，除非另有说明。
In Eq.~\ref{lif2}, $\Delta t$ is the time step, $v_{reset}$ is the reset potential, $u(t)$ and $v(t)$ represent pre- and post-spike membrane potentials, and $g(\cdot)$ is the Heaviside function modeling spiking behavior. After the loss of the reward signal, updating the network weights stops. This strategy of directly optimizing the weights is set as a control group in our experiments. Neuronal parameters are given in Tab.~\ref{params} unless otherwise specified.

% traces是突触前或突触后神经元的尖峰触发在突触前和突触后部位产生的轨迹。一般来说，这些迹线代表突触前后神经元最近的激活水平~\cite{pfister2006triplets}。迹可以通过在模型中使用线性算子和在电路中使用低通滤波器对尖峰进行积分来计算，或者通过使用非线性算子/电路来计算。在实验中, 突触的迹被建模为: 
\emph{Traces} are the tracks produced at the pre- and post-synaptic sites by the spikes of pre- or post-synaptic neurons. Generally, these traces represent the recent activation level of pre- and post-synaptic neurons~\cite{pfister2006triplets}. Traces can be computed by integrating spikes using a linear operator in the model and a low-pass filter in the circuit or by using non-linear operators/circuits. In the experiments, the synaptic traces were modeled as follows:

\begin{equation}
    x(t) = \sum_{\tau=0}^{t} \lambda^{t-\tau}s(\tau)
    \label{trace}
\end{equation}

% 在这个等式中，$x_{t}$是时间$t$的突触轨迹，$\lambda$是衰减因子，反映了尖峰的影响随时间的消退速度，$s(\tau)$表示时间$\tau$的尖峰。如式子xxx所示，这些突触踪迹用于保持神经元激活的短期历史，并与元可塑性结合, 调控突触权重. 
In Eq.~\ref{trace}, $x(t)$ is the synaptic trace at time $t$, $\lambda$ is the decay factor reflecting how quickly a spike's influence fades with time, and $s(\tau)$ represents the spike at time $\tau$. In the context of our experiment, these synaptic traces maintain a short-term history of neuron activation, thereby adding an element of temporal dynamics to our network model. As shown in Eq.~\ref{plasticity_centric}, these synaptic traces are used to maintain a short-term history of neuronal activation and, in conjunction with PDLF, to modulate synaptic weights.

\begin{table*}[htbp]
    \centering
    \caption{Parameters of the spiking neurons.}
    \label{params}
    \begin{tabular}{cccc}
        \toprule
        \multirow{2}{*}{\bf Parameter} & \multicolumn{2}{c}{\bf Value} & \multirow{2}{*}{\bf Description}                          \\
        \cmidrule(lr){2-3}
                                       & WM task                       & RL task                          &                        \\
        \hline
        $\Delta t$                     & $20$ ms                       & $200$ ms                         & Simulation time step   \\
        $\tau_m$                       & $40$ ms                       & $400$ ms                         & Membrane time constant \\
        $\lambda$                      & $54$ ms                       & $544$ ms                         & Decay factor           \\
        $v_{th}$                       & $0.1$ V                       & $0.1$ V                          & Membrane threshold     \\
        $v_{reset}$                    & $0$ mV                        & $0$ mV                           & Reset potential        \\
        $v_{0}$                        & $0$ mV                        & $0$ mV                           & Resting potential      \\
        \bottomrule
    \end{tabular}
\end{table*}

\subsection{Experimental Settings}

\subsubsection{Working Memory Task}

% 为了验证元可塑性对于工作记忆的影响, 我们设计了工作记忆任务. 智能体会先接收到一个刺激序列, 并在$m$步的延迟之后, 智能体被要求重复收到的刺激. 每次实验中, 一个长度为$n$的随机序列会被生成, 其中$r_t \sim \mathcal{B}(1, \frac{1}{2}), 1 < t \leq n$.  在每一个时间步, 输入是一个三维的向量$\vec{a_t}$, 并可以分为三个阶段:
To validate the impact of PDLF on working memory, we designed a working memory task. The agent would first receive a stimulus sequence, and after a delay of $m$ steps, the agent is asked to reproduce the received stimulus. In each experiment, a random sequence of length $n$ would be generated, where $r_t \sim \mathcal{B}(1, \frac{1}{2}), 1 < t \leq n$. At each time step, the input is a three-dimensional vector $\vec{a_t}$, which can be divided into three stages:

% 刺激接收： 如果$1 < t \leq n$，$\vec{a_t} = (r_t, 1, 0)$。第一个元素是输入刺激的类型，第二个元素是输入刺激的指示符。
% 延迟 如果$n + 1 < t \leq n + m$，$\vec{a_t} = (0, 0, 0)$。该阶段代表一个没有新刺激出现的延迟期。
% 刺激再现： 如果 $n + m + 1 < t \leq 2n + m$，$\vec{a_t} = (0, 0, 1)$。最后一个元素表示脉冲是否需要重现。

\begin{itemize}
    \item Stimulus reception: If $1 < t \leq n$, $\vec{a_t} = (r_t, 1, 0)$. The first element is the type of input stimulus, and the second element is the indicator for the input stimulus.
    \item Delay: If $n + 1 < t \leq n + m$, $\vec{a_t} = (0, 0, 0)$. This phase represents a delay period where no new stimulus is presented.
    \item Stimulus reproduction: If $n + m + 1 < t \leq 2n + m$, $\vec{a_t} = (0, 0, 1)$. The last element indicates whether a pulse needs to be reproduced.
\end{itemize}

% 在每一步骤, 模型会有一个标量的输出 $s_t$, 是对于刺激的预测. 最后$m$步的均方误差损失被作为模型的奖励:
At each step, the model has a scalar output $s_t$, which is a prediction for the stimulus. The Mean Square Error over the last $m$ steps is taken as the reward of the model:

\begin{equation}
    R = -\frac{1}{n} \sum_{\tau = 1}^{n} (r_t - s_{m + n + \tau})^2
    \label{memory_reward}
\end{equation}

% 式xxx中的MSE被作为训练中智能体的奖励函数, 为了直观地对比具有不同策略的智能体, 我们使用每一步的平均正确率作为在测试中, 衡量模型性能的准则: 
Eq.~\ref{memory_reward} is used as a reward function in training. To intuitively compare agents with different strategies, as shown in Fig.~\ref{memory}\textbf{B}, we utilize the average accuracy per step as the performance measure during testing, as shown in Eq.~\ref{memory_acc}.

\begin{equation}
    \text{Acc} = \frac{1}{n} \sum_{\tau = 1}^{n} (r_t == s_{m + n + \tau})
    \label{memory_acc}
\end{equation}

% 在接收刺激阶段, 每一个刺激服从分布$\mathcal{B}(1, \frac{1}{2})$, 也就是说, 机会水平的平均准确率为 $0.5$. 
During the stimulus reception stage, each stimulus follows the distribution $\mathcal{B}(1, \frac{1}{2})$, which means that the average accuracy at the chance level is $0.5$.

\subsubsection{Multi-task Reinforcement Learning}

% 我们在基于Brax模拟器的五个连续控制环境（ant\_dir, swimmer\_dir, halfcheetah\vel, hopper\_vel, ur5e, fetch）上评估了我们的方法. 

% ant\_dir： 在该环境中，我们训练蚂蚁代理朝一个目标方向运行。训练任务集包含8个方向, 从[0,360]度均匀采样。如图~\ref{rl_exp}\textbf{D}所示, 泛化性测试任务集包含72个方向，从[0,360]度均匀采样。Agent的奖励有沿目标方向以速度，和控制开销组成。

% swimmer\_dir: 在该环境中, 我们训练一个游泳者朝着一个固定的方向移动. 训练和测试任务的设置与ant\_dir相似. 

% halfcheetah\_vel: 在HalfCheetah_Vel环境中，我们训练半身猎豹智能体以特定的速度前进。训练任务包括8个速度, 从[1, 10] m/s 中均匀采样. 泛化性测试任务包含72个不同的速度, 从与训练任务相同的范围中均匀采样. 

% hopper\_vel: 在hopper\_vel环境中, 我们训练蚂蚱智能体学会以特定的速度前进。实验设置与halfcheetah\_vel相同, 但是速度的采样区间为[0, 2] m/s. 

% ur5e: UR5e是一种常见的6自由度机械臂，常用于工业自动化和机器人研究领域。在机械臂的末端和目标位置之间的距离小于 0.02m 时, 会获得奖励. 随后目标位置会被随机重置. 智能体的目标是在规定时间内, 尽可能多的达到目标位置. 

% fetch: 在该环境中, 我们训练一个狗智能体跑向目标地点. 实验设置与ur5e相同. 

% 智能体在所有的任务上的平均奖励被作为最终的奖励, 以训练智能体能够同时学会多个任务. 

We evaluated our method on five continuous control environments based on the Brax simulator (\texttt{ant\_dir}, \texttt{swimmer\_dir}, \texttt{halfcheetah\_vel}, \texttt{hopper\_vel}, \texttt{ur5e}, \texttt{fetch}).

\begin{itemize}
    \item \texttt{ant\_dir}: We train an ant agent to run in a target direction in this environment. The training task set includes $8$ directions, uniformly sampled from $[0,360]$ degrees. As shown in Fig.~\ref{rl_exp}\textbf{D}, the generalization test task set includes $72$ directions, uniformly sampled from $[0,360]$ degrees. The agent's reward comprises speed along the target direction and control cost.

    \item \texttt{swimmer\_dir}: In this environment, we train a swimmer agent to move in a fixed direction. The settings for training and testing tasks are similar to \texttt{ant\_dir}.

    \item \texttt{halfcheetah\_vel}: In the \texttt{halfCheetah\_vel} environment, we train a half-cheetah agent to move forward at a specific speed. The training tasks include $8$ speeds, uniformly sampled from [1, 10] m/s. The generalization test tasks include $72$ different speeds, uniformly sampled from the same range as the training tasks.

    \item \texttt{hopper\_vel}: In the \texttt{hopper\_vel} environment, we train a hopper agent to advance at a specific speed. The experimental setup is the same as \texttt{halfcheetah\_vel}, but the sampling interval for the speed is $[0, 2]$ m/s.

    \item \texttt{ur5e}: The UR5e is a common 6-DOF (degrees of freedom) robotic arm frequently used in industrial automation and robotics research. The agent receives a reward when the distance between the robotic arm's end and the target position is less than $0.02$ m. The target position is then randomly reset. The agent's goal is to reach the target position as many times as possible within the stipulated time.

    \item \texttt{fetch}: We train a dog agent to run to a target location in this environment. The experimental setup is similar to \texttt{ur5e}.
\end{itemize}

The agent's final reward is the average reward across all tasks, which encourages the agent to learn multiple tasks simultaneously.

\subsection{Training Strategies}

\begin{algorithm}
    \caption{Parameter-Exploring Policy Gradients (PEPG)}
    \label{alg:PEPG}
    \begin{algorithmic}[1]
        % 初始化代数为$M$, 种群数量为$N$
        \State Initialize the number of generations $M$, and the population size $N$
        \State Initialize policy parameters $\theta$
        \State Initialize adaptive noise scaling parameters $\sigma$
        \State Initialize learning rates $\alpha_\theta$, $\alpha_\sigma$
        \State Initialize Adam parameters $m_\theta, v_\theta, m_\sigma, v_\sigma, \beta_1, \beta_2, \epsilon$
        \State Initialize noise standard deviation $\sigma_0$
        \For{$m=1$ to $M$}
        \For{$n=1$ to $N$}
        \State Sample noise $\epsilon \sim \mathcal{N}(0, \sigma_0)$
        \State Compute offspring $\theta' = \theta + \sigma \odot \epsilon$
        \State Evaluate fitness $f(\theta')$
        \EndFor
        \State Compute fitness baseline $b = mean(f(\theta'))$
        \State Compute gradients $\nabla_\theta = \frac{1}{N}\sum_{i=1}^{N} f_i \cdot \epsilon_i$
        \State Compute adaptive noise scaling gradient $\nabla_\sigma = \frac{1}{2N} \sum_{i=1}^{N} ((f_i - b)^2 - \sigma^2)$
        \State Update Adam parameters for $\theta$: $m_\theta = \beta_1 m_\theta + (1 - \beta_1) \nabla_\theta$, $v_\theta = \beta_2 v_\theta + (1 - \beta_2) \nabla_\theta^2$
        \State Update policy parameters $\theta = \theta + \alpha_\theta \cdot \frac{m_\theta}{\sqrt{v_\theta} + \epsilon}$
        \State Update Adam parameters for $\sigma$: $m_\sigma = \beta_1 m_\sigma + (1 - \beta_1) \nabla_\sigma$, $v_\sigma = \beta_2 v_\sigma + (1 - \beta_2) \nabla_\sigma^2$
        \State Update adaptive noise scaling $\sigma = \sigma \exp(\alpha_\sigma \cdot \frac{m_\sigma}{\sqrt{v_\sigma} + \epsilon})$
        \EndFor
    \end{algorithmic}
\end{algorithm}

\begin{table}[ht]
    \caption{Parameters in PEPG.}
    \label{table:pepg_param}
    \centering
    \begin{tabular}{lll}
        \hline
        \bf Parameter        & \bf Value    & \bf Description                                    \\
        \hline
        $\theta$             & $0$          & Initial policy parameters                          \\
        $\sigma$             & $0.1$        & Initial adaptive noise scaling parameters          \\
        $\alpha_\theta$      & $0.15$       & Learning rate for policy parameters                \\
        $\alpha_\sigma$      & $0.1$        & Learning rate for adaptive noise scaling           \\
        $m_\theta, v_\theta$ & $0, 0$       & Initial Adam parameters for policy parameters      \\
        $m_\sigma, v_\sigma$ & $0, 0$       & Initial Adam parameters for adaptive noise scaling \\
        $\beta_1, \beta_2$   & $0.9, 0.999$ & Hyperparameters of Adam optimizer                  \\
        $\epsilon$           & $10^{-8}$    & Adam parameters                                    \\
        $M$                  & $1500$       & Number of generations                              \\
        $N$                  & $128$        & Number of offspring per generation                 \\
        \hline
    \end{tabular}
\end{table}

% 我们使用演化策略优化SNNs. 对于具有可塑性的SNNs而言, Eq.~\ref{plasticity_centric}中的可塑性的参数被用于优化. 不同代之间, 通过更改突触之间的可塑性规则进行演化, 而不是直接地修改权重. 权重直接训练的SNNs被作为对照组, 突触的权重被作为优化参数. 实验中使用的PEPG由算法~\ref{alg:PEPG}提供, 如无特殊说明, 具体的参数设置以及对于不同参数的解释如表~\ref{table:pepg_param}所示. 计算适应度$f(\theta)$的方式根据任务不同而改变, 对于工作记忆任务, 适应度由Eq.~\ref{memory_reward}提供, 而对于多任务强化学习而言, 适应度是不同子任务的平均回合奖励. 
We employ Parameter-Exploring Policy Gradients (PEPG)~\cite{sehnke2010parameter} to optimize SNNs. For SNNs with plasticity, the plasticity parameters in Eq.~\ref{plasticity_centric} are used for optimization. Evolution across generations is facilitated by modifying synaptic plasticity rules rather than directly adjusting the weights. SNNs with directly trained weights are considered a control group, where synaptic weights are the optimization parameters. The implementation of PEPG used in the experiments is provided by Algorithm~\ref{alg:PEPG}. Unless expressly stated otherwise, the parameter settings and their explanations are shown in Table~\ref{table:pepg_param}. The way to compute fitness $f(\theta)$ varies depending on the task. For the working memory task, fitness is provided by Eq.~\ref{memory_reward}, while for multi-task reinforcement learning, fitness is the average episodic reward across different subtasks.

% \section*{Acknowledgement}
% This work was supported by the National Key Research and Development Program (Grant No. 2020AAA0104305), and the Strategic Priority Research Program of the Chinese Academy of Sciences (Grant No. XDB32070100).

\section*{Data Availability}
Codes and data have been deposited in GitHub https://github.com/FloyedShen/PDLF~\cite{metaplastic2023github}.

\bibliography{refs}
\bibliographystyle{unsrt}

\end{document}